\documentclass{article}
\usepackage{spconf,amsmath,graphicx}
 \usepackage{booktabs,url,multirow}
 \usepackage{amssymb}
 \usepackage{enumitem}

\title{Dialect Adaptation and Data Augmentation for Low-Resource ASR: TalTech Systems for the MADASR 2023 Challenge}
\name{Tanel Alumäe, Jiaming Kong, Daniil Robnikov}
\address{Department of Software Science\\
Tallinn University of Technology, Estonia}
\copyrightnotice{979-8-3503-0689-7/23/\$31.00~\copyright2023 IEEE}
\begin{document}
%
\maketitle
\begin{abstract}
This paper describes Tallinn University of Technology (TalTech) systems developed for the ASRU MADASR 2023 Challenge. The challenge focuses on automatic speech recognition of dialect-rich Indian languages with limited training audio and text data.  TalTech participated in two tracks of the challenge: Track 1 that allowed using only the provided training data and Track 3 which allowed using additional audio data. In both tracks, we relied on wav2vec2.0 models. Our methodology diverges from the traditional procedure of finetuning pretrained wav2vec2.0 models in two key points: firstly, through the implementation of the aligned data augmentation technique to enhance the linguistic diversity of the training data, and secondly, via the application of deep prefix tuning for dialect adaptation of wav2vec2.0 models. In both tracks, our approach yielded significant improvements over the provided baselines, achieving the lowest word error rates across all participating teams.
\end{abstract}
\section{Introduction}
\label{sec:intro}

The ASRU 2023 MADASR (\textit{Model Adaptation for ASR in Low-Resource Indian Languages}) Challenge\footnote{\url{https://sites.google.com/view/respinasrchallenge2023/home}} \cite{singh2023model} focused on automatic speech recognition (ASR) of dialect-rich Indian languages with limited training audio and text data. The aim of the challenge was to investigate the importance of acoustic and text-based training data, experiment with different model architectures and modeling techniques and validate the importance of various pretrained models. Evaluation was done on two languages (Bengali and Bhojpuri) across four individual tracks, each permitting a varying degree of external acoustic and textual data:
\begin{itemize}[noitemsep,topsep=0pt,parsep=5pt,partopsep=2pt]
    \item Track 1 restricts participants to use only the provided training corpus for both acoustic and language modeling;
    \item Track 2 allows using additional text data;
    \item Track 3 allows using additional acoustic data (including pretrained models), but doesn't allow using additional textual data;
    \item Track 4 encourages participants to use any existing resources along with the provided  audio and text data.
\end{itemize}
Furthermore, this challenge assessed the development of ASR systems for two languages under considerable time constraints, as the duration allocated for training the models was approximately one month.  Word error rate (WER) and character error rate (CER) were used as evaluation criteria.

This paper describes the systems developed at Tallinn University of Technology (TalTech) for the challenge. We participated in Tracks 1 and 3. Our main focus was investigating various approaches of pretraining end-to-end ASR models. We also experimented with dialect-adaptive ASR models and tackled the low linguistic variety of the provided training data by implementing various augmentation techniques. Notably, our approach deviates from the conventional procedure of fine-tuning  pretrained wav2vec2.0 models in two key aspects: the adoption of the aligned word replacement augmentation technique to enhance the linguistic diversity of the training data and the use of deep prefix tuning for dialect adaptation of wav2vec2.0 models. According to our knowledge, this is the first work showing that prefix tuning can be efficiently used for adapting wav2vec2.0 based models. In the post-evaluation phase, we explored text-to-speech based data augmentation, obtaining further improvements.

\section{Data}
\label{sec:data}
\begin{table}[tb]
\caption{Statistics of the provided speech data.}
\label{tab:data}
\centering
\begin{tabular}{@{}l|r|r@{}}
\toprule
                   & Bengali & Bhojpuri \\ \midrule
\multicolumn{3}{l}{\textit{Speech data}} \\ \midrule
\#Hours            & 852     & 835      \\
\#Speakers         & 1980    & 1926     \\
\#Utterances       & 579996  & 625709   \\
\#Unique sentences & 17034   & 19620    \\ 
\#Dialects         & 5       & 3       \\
\#Dev. utterances  &  1240  & 1262 \\
Dev. OOV rate    & 4.2\% & 2.8\% \\
\#Test utterances & 3290 & 3010 \\ \midrule
\multicolumn{3}{l}{\textit{Additional textual data}} \\ \midrule
\#Sentences & 194K & 228K \\ 
\#Words & 2.6M & 3.1M \\
\bottomrule
\end{tabular}
\end{table}

\begin{figure}[tb]
  \centering
  \includegraphics[width=0.99\linewidth]{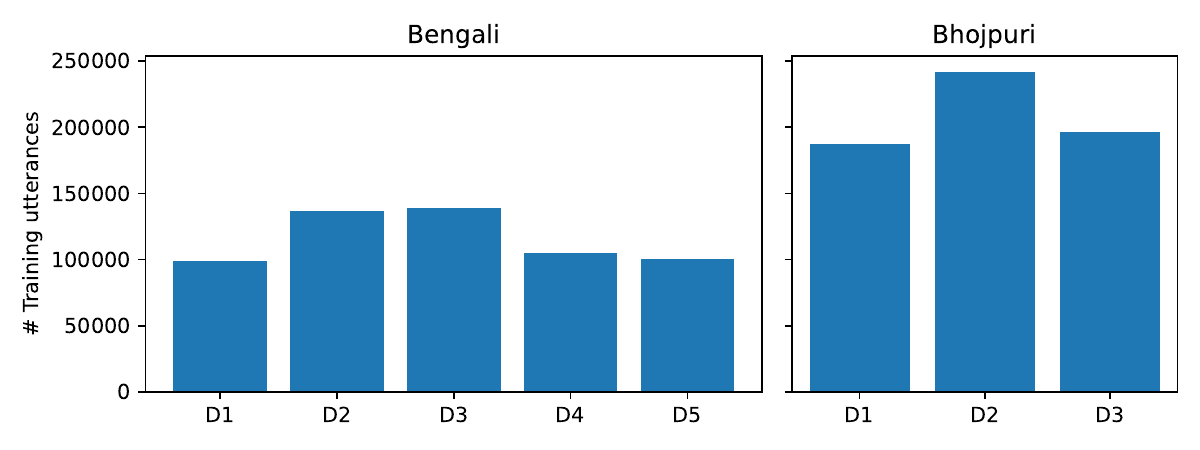}
  \caption{Distribution of different dialects in the training data.}
  \label{fig:dialects}
\end{figure}

The statistics of the training and development data provided by the challenge organizers is described in Table \ref{tab:data}. The speech data was collected within the RESPIN project, which is an initiative aimed to collect dialect-rich speech corpora in 9 Indian languages. The training datasets contain read speech, with occasional background noise. Metadata about the speaker's identity and dialect background is provided for each utterance in the training, development, and test data. The frequency distribution of different dialects in the training data is shown in Figure \ref{fig:dialects}. 

One notable aspect of the provided training data is the relatively low linguistic variety of the training utterances: e.g., for Bengali, there are almost 580K training utterances but only 17K unique sentences, meaning that each sentence repeats approximately 34 times in the training data, spoken by different speakers. 

In addition to the speech data, the challenge organizers provided additional textual data for both languages, as shown in Table \ref{tab:data}. The textual data is categorized according to the dialect and domain (such as healthcare, politics, etc).

\section{Track 1 System}

\subsection{Conformer Wav2vec2.0 Pretraining}
\label{sec:conformer_w2v}

A wav2vec2.0 model \cite{wav2vec2} is trained on unlabeled speech data by jointly solving a contrastive task over masked latent speech representations and learning a quantization of the latents.  
The model contains a convolutional feature encoder that maps raw audio to latent speech representations, which are fed to a Transformer network that outputs context representations.

Although wav2vec2.0 based pretraining is especially useful in cases when large amounts of unlabeled data is available for self-supervised training, it has been shown that ASR model can benefit from a pretraining step even when only the supervised dataset is used for pretraining \cite{wav2vec2}. Therefore, since no additional speech data was allowed to be used in Track 1, we pretrained separate Conformer-based wav2vec2.0 models for both languages on the provided speech data and then fine-tuned them for ASR using the character-based CTC objective on the same data. 

Wav2vec2.0-based pretraining and fine-tuning was implemented using \textit{fairseq} \cite{fairseq}. We used the ``conformer-base'' architecture that contains 12 Conformer blocks, model dimension 768, inner dimension
 3,072 and 12 attention heads. Pretraining was done using an effective batch size of around 107 minutes, for 100,000 iterations. This took  approximately three days per language, using four A100 GPUs with 80 GB GPU RAM. We used
 a peak learning rate of 0.00005 (with a warm-up period of 16000 iterations). The rest of the pretraining hyperparameters were taken from the \textit{fairseq}  Librispeech ``conformer-base''  recipe\footnote{\url{https://github.com/facebookresearch/fairseq/blob/main/examples/wav2vec/config/pretraining/wav2vec2_conformer_base_librispeech.yaml}}. During the fine-tuning process, the CTC objective was employed, using a character-based output vocabulary. An effective batch size of 320 seconds was used, accompanied by a peak learning rate of 0.00001. The learning rate was gradually increased and subsequently decreased according to a linear schedule. Initially, only the output classifier was trained for the first 10,000 updates. Following this phase, the Conformer layers were also fine-tuned. Throughout the fine-tuning process, the convolutional feature encoder remained frozen and unchanged. Additionally, SpecAugment was applied to the features produced by the feature encoder, employing a timestep mask probability of 0.5 and a channel mask probability of 0.1. The channel mask length used was 64. Furthermore, LayerDrop, with a probability of 0.1, was applied as an additional regularization technique.

\begin{table}[tb]
\caption{Conformer WER results on the Bengali dev set with and without wav2vec2.0 self-supervised pretraining.}
\label{tab:wav2vec}
\centering
\begin{tabular}{l|c}
\toprule
& WER \\ \midrule
Without pretraining & 25.1 \\
With pretraining & 23.0 \\
\bottomrule
\end{tabular}
\end{table}
The comparison of Word Error Rate (WER) results on the Bengali development set, with and without pretraining, is presented in Table \ref{tab:wav2vec}. The non-pretrained model employs an architecture that mirrors the wav2vec2.0 Conformer-based model, but undergoes supervised training starting from scratch. In this model, the learning rate was independently tuned, and the feature encoder was trained in conjunction with the remaining model parameters. The results indicate that pretraining leads to an approximately 10\% relative reduction in WER compared to the non-pretrained model.

\subsection{Dialect Adaptation via Prefix Tuning}

As explained earlier, all training, development and test utterances are categorized according to the spoken dialect. The number of dialects is fixed for both languages. This leads to the idea of adapting end-to-end ASR models trained on all language data for each dialect. The most straightforward method would be to fine-tune the dialect-independent model on each of the dialect subsets, thereby generating distinct models tailored to each dialect. However, this approach necessitates the storage and loading of dialect-specific models whenever required for processing individual utterances, resulting in substantial computational overhead.

An alternative approach involves using a method categorized as parameter-efficient fine-tuning (PEFT). PEFT methods allow for the efficient adaptation of pretrained neural models to different use cases without requiring the fine-tuning of all model parameters. In this regard, PEFT methods only fine-tune a small number of (possibly additional) model parameters, significantly reducing computational and storage costs. Recent state-of-the-art PEFT techniques, such as LoRA \cite{hu2021lora} and prompt/prefix tuning \cite{li2021prefixtuning, liu-etal-2022-p,liu2021gpt}, achieve comparable performance to full fine-tuning.

\begin{figure}[tb]
  \centering
  \includegraphics[width=0.99\linewidth]{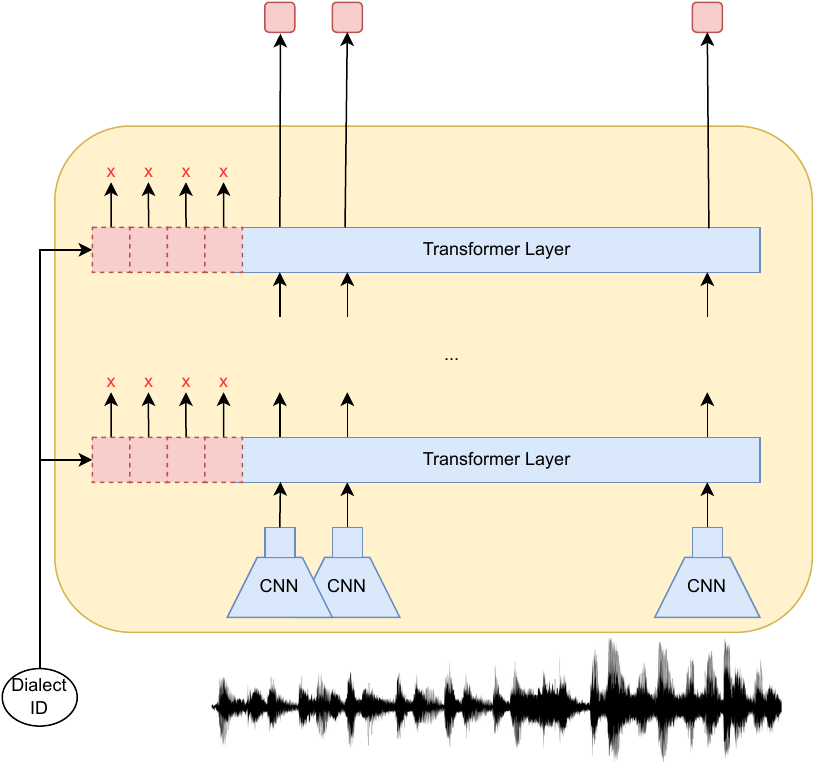}
  \caption{Dialect-based prefix tuning (with prefix length 4) of a Transformer CTC model.}
  \label{fig:prefix-tuning}
\end{figure}

We chose to adapt dialect-agnostic models to different dialects using prefix-tuning \cite{li2021prefixtuning}, as described on Figure \ref{fig:prefix-tuning}.  For
each dialect $d$, we inject $L_{prefix}$ prefix vectors into the model, parameterized by $\theta_p$. 
As proposed in \cite{li2021prefixtuning}, the prefix vectors are prepended to each Transformer/Conformer layer as additional key and value vectors. Positional encoding is not applied to the prefix vectors.
The output of the Transformer layer that corresponds to the prefix is discarded. The dimensionality of $\theta_p$ is thus $L_{prefix} \times \text{\#Layers}  \times \text{hidden dim} \times 2$. When using a large MMS wav2vec2.0 model with the hidden dimensionality of 1280 and 48 layers, and setting the prefix length to 4 (that we used in this challenge), then this translates to only about 500K additional parameters per dialect.

The backbone dialect-independent model can be trained jointly with the prefix vectors, or it can be pretrained in a dialect-independent manner. Based on our initial experiments conducted during the early stages of this challenge, we observed no substantial difference between these two approaches. Due to technical considerations, we ultimately decided to first pretrain a dialect-agnostic model and then proceed with learning the prefix vectors. The prefix vectors are trained using an effective batch size of 192 utterances, with a learning rate of 0.01 and weight decay of 0.001. We observed that only 1-2 epochs is needed for training the prefixes.

Table \ref{tab:prefix-tuning} presents the  WER results for both languages and two wav2vec2.0 models, comparing the performance with and without dialect-based prefix-tuning. The proposed technique consistently leads to improvements in WER, with the improvement slightly reduced when shallow fusion with a dialect-agnostic 4-gram language model is used during decoding. Dialect adaptation is more effective for Bengali, which has five dialects.

\begin{table}[tb]
\caption{Dialect adaptation via prefix tuning improves WER over both languages and across all models.}
\label{tab:prefix-tuning}
\begin{tabular}{l|p{1cm}|cp{0.7cm}|cp{0.7cm}}
\toprule
                           &                       & \multicolumn{2}{c|}{Bengali} & \multicolumn{2}{c}{Bhojpuri} \\ \midrule
                           
Model                      & Dialect prefix tuning & No LM        & With LM       & No LM        & With LM        \\ \midrule
\multirow{2}{*}{Conformer} &                     & 21.3         & 19.7         & 19.5         & 18.1          \\
                           & \hfil \checkmark                   & 20.0         & 18.3         & 19.3         & 17.8          \\ \midrule
\multirow{2}{*}{MMS-1B}                     &                     & 17.6         & 16.9         & 16.8         & 16.3          \\
                           & \hfil \checkmark                   & 16.7         & 16.6         & 15.8         & 15.9 \\ \bottomrule        
\end{tabular}
\end{table}

\subsection{Aligned  Data Augmentation}

In the analysis presented in Section \ref{sec:data}, it was shown that the training data features a considerable amount of repeating prompts. For instance, out of the 580K training utterances for Bengali, merely 17K sentences are unique. It swiftly became evident that this redundancy causes significant overfitting problems, predominantly with model architectures that inherently incorporate a language model component, such as sequence-to-sequence models and neural transducers. Even though the impact on CTC-based models isn't as pronounced, it remains clear that such models also tend to overfit more rapidly on this data than anticipated.

To enrich the linguistic diversity of the training data, we implemented a strategy known as Aligned Data Augmentation (ADA) \cite{lam21b_interspeech}. This technique exploits audio and text alignment data to generate new training examples, by substituting a given percentage of words in both the transcripts and the audio with arbitrary words from other utterances. The technique requires the preliminary training of a model that can be used for aligning transcripts with the corresponding audio. Despite the original research utilizing a DNN-HMM based model for this purpose, we chose to employ a CTC-based wav2vec2.0 Conformer model that we already possessed for both languages. CTC-based models have demonstrated high accuracy in generating alignment information \cite{ctcsegmentation}.

In contrast to the original proposal, we conducted the augmentation offline according to the following protocol: approximately 20\% of the words within all training sentences and their corresponding utterances were substituted with arbitrary words from other utterances made by the same speaker. Even though this constraint of employing utterances from the same speaker was not specified in the initial method proposal, we observed that this adjustment resulted in the augmented utterances sounding significantly more natural.

\begin{figure}[tb]
  \centering
  \includegraphics[width=0.99\linewidth]{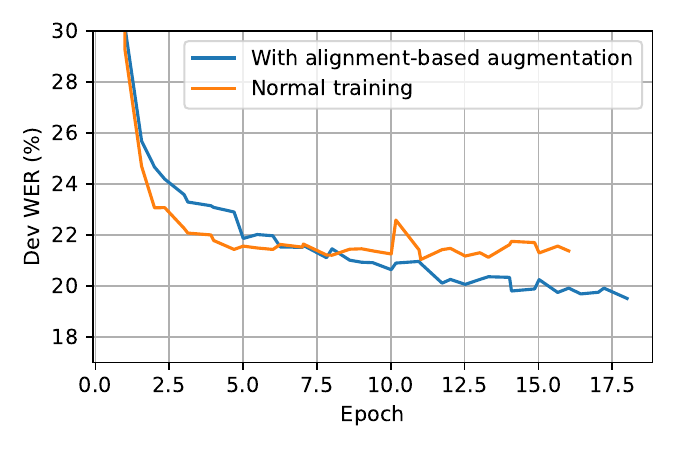}
  \caption{WER of development data during normal training \textit{vs} alignment-based augmentation.}
  \label{fig:ada}
\end{figure}

Figure \ref{fig:ada} illustrates the WER curves for the Bengali development data, comparing standard training against training with alignment-based augmentation with Conformer-based wav2vec2.0 models described before. Although convergence appears slower on augmented data, it effectively circumvents the overfitting issue observed in the standard training and results in a relative reduction of more than 10\% in WER.

\subsection{Language Modelling}
\label{sec:lm}

As outlined in Section \ref{sec:data}, the number of unique sentences in the training data is relatively low, falling below 20K for both languages. The challenge provides additional textual training data that can be used in all tracks. We conducted several experiments to find the best way to combine the data into language model (LM). Due to time constraints, we only considered n-gram models.

Table \ref{tab:data} shows that an LM trained only on the speech transcripts has a relatively large out-of-vocabulary (OOV) rate -- 4.2\% and 2.8\% for Bengali and Bhojpuri, respectively. Using all words from the additional textual training data decreases the OOV rate to 1.0\% for Bengali and 1.1\% for Bhojpuri.

Table \ref{tab:lm} presents the LM perplexities associated with three distinct 4-gram models, along with the corresponding WERs when employing the Conformer wav2vec2.0 model (trained with ADA and dialect adaptation) for decoding. All 4-gram models used Kneser-Ney (KN) smoothing with lower-order interpolation, and all n-grams appearing in the training data were incorporated into the model. The final model (interpolated mixture) was created by training a single model on all text domains, including speech transcripts, followed by interpolating the model into the final one using weights optimized on development data. The SRILM toolkit \cite{stolcke2011srilm} was utilized for LM training.

The table demonstrates that the use of LMs (especially the ones trained on additional data) improves WER by a large margin, with relative reduction reaching up to 25\% for Bengali. The LM trained across all training data yields the lowest WER for both languages, despite its slightly higher perplexity compared to the interpolated model. However, the interpolated model was still chosen for our submission.

\begin{table}[]
\caption{LM perplexities and shallow fusion WERs, when using the Conformer wav2vec2.0 model for decoding.}
\label{tab:lm}
\begin{tabular}{p{3.8cm}|rr|rr}
\toprule
                                         & \multicolumn{2}{c|}{Bengali} & \multicolumn{2}{c}{Bhojpuri} \\ \midrule
Language model                                         & PPL           & WER         & PPL           & WER          \\ \midrule
-                                    &               & 20.0        &               & 19.3         \\
KN 4-gram, trained on speech transcripts & 185         & 18.3        & 126         & 17.7         \\
KN 4-gram, with additional data          & 88          & 14.7        & 69          & 15.1         \\
KN 4-gram, interpolated mixture          & 83          & 15.0        & 64          & 15.3    \\ \bottomrule    
\end{tabular}
\end{table}

\subsection{System Combination}

Our system for producing the final submission was based on multiple models. Each of the models was used for generating N-best list of ASR hypotheses.
We adopted the N-best list combination method  that was successfully used in \cite{alumae2021combining}, consisting of the following steps and implemented using SRILM:
\begin{enumerate}[noitemsep,topsep=0pt,parsep=5pt,partopsep=2pt]
    \item N-best (we used $N=50$) lists are generated for all subsystems, including acoustic and language model scores for each hypothesis;
    \item Optimized weights for acoustic and language model scores and sentence length normalization factors are found for each subsystem separately, using  simplex-based ``Amoeba'' search of the lowest WER on development data;
    \item N-best list combination weights are optimized on the development data, using the resulting posteriors of reference words as the optimization target;
    \item N-best lists with optimized score combination weights are decoded using word error minimization, in a generalization of the ROVER algorithm \cite{stolcke1997explicit}.
\end{enumerate}

\subsection{Final system}

In Track 1, our final system was based on two models: (1) a Conformer wav2vec2.0 model described in Section \ref{sec:conformer_w2v}, trained with ADA, and (2) the same model adapted to dialects using prefix tuning. For both target languages, we used the same model architectures and training hyperparameters. WER results of the official baseline model (codenamed espnet-conformer), our individual models and their fusion are presented in Table \ref{tab:track1-results}.  Note that the WER results in Table \ref{tab:lm}  are produced using beam search while those
in Table \ref{tab:track1-results} are the result of confusion decoding of N-best lists. Thus, those numbers
are not directly comparable.
Our submission resulted in the lowest WER scores among all Track 1 participants.

\begin{table}[]
\caption{Track 1 WERs of our individual systems (after optimizing LM weight) and their fusion, along with the official challenge baseline WERs.}
\label{tab:track1-results}
\begin{tabular}{l|cc|cc}
\toprule
                              & \multicolumn{2}{c|}{Bengali} & \multicolumn{2}{c}{Bhojpuri} \\ \midrule
                              & Dev          & Test         & Dev           & Test         \\ \midrule 
Challenge baseline & 20.7 & 22.1 & 20.1 & 20.2 \\ \midrule        \midrule                      
Conformer wav2vec2.0          & 15.9         &              & 15.5          &              \\
+ dialect-based prefix tuning & 15.2         &              & 15.3          &              \\ \midrule
\textbf{Fusion}                        & \textbf{14.9}         & \textbf{20.2}    &     \textbf{15.2}          & \textbf{16.8}        \\ \bottomrule
\end{tabular}
\end{table}

\section{Track 3 System}

\subsection{Zero-shot results}

\begin{table}[tb]
\caption{Zero-shot Bengali WER results with two large-scale multilingual models.}
\label{tab:zero-shot}
\centering
\begin{tabular}{l|c}
\toprule
                  & Bengali WER \\ \midrule
MMS-1B-all (with Bengali adapter) & 49.5    \\
Whisper-large-v2       & 112             \\ \bottomrule
\end{tabular}
\end{table}

Recently, two large-scale publicly available multilingual ASR models have been released that can also process Bengali speech. However, neither can decode Bhojpuri. Table \ref{tab:zero-shot} compares the out-of-the-box performance on Bengali development data of the large Whisper model \cite{radford2022whisper} and the MMS-1B wav2vec2.0 model fine-tuned for multilingual ASR by the authors \cite{pratap2023scaling}. While the MMS model exhibits promising performance, the Whisper model fails on the challenge data. This is the main reason why we focused our efforts in Track 3 on fine-tuning wav2vec2.0 models. 

Our initial investigations also included the fine-tuning of Whisper models, resulting in WERs comparable to those achieved with wav2vec2.0-based methods. However, due to the significantly higher computational demands associated with fine-tuning and decoding using Whisper's sequence-to-sequence models, coupled with the challenge's time restrictions, we opted to reject this approach.

\subsection{Comparison of backbone wav2vec2.0 models}

\begin{table}[tb]
\caption{Comparison of wav2vec2.0 models after finetuning on Bengali.}
\centering
\label{tab:backbone-comp}
\begin{tabular}{l|c}
\toprule
                  & Bengali WER \\ \midrule
Track 1 Conformer & 20.7        \\
XLS-R-1B          & 18.1        \\
CLSRIL-23         & 18.0        \\
MMS-1B            & 17.4       \\ 
MMS-1B-all        & 17.4        \\
\bottomrule
\end{tabular}
\end{table}

The multilingually pretrained wav2vec2.0 models have demonstrated their effectiveness in low-resource ASR \cite{babu2021xlsr,pratap2023scaling}. As an initial investigation, we sought to determine which of the backbone models yields the most optimal results after fine-tuning on the provided data. Due to time restrictions, we focused our benchmarking solely on Bengali data. We compared our Conformer wav2vec2.0 model from Track 1,  XLS-R-1B \cite{babu2021xlsr}, CLSRIL-23 (pretrained on 10K hours of 23 Indian languages) \cite{gupta2022clsril23}, and  MMS-1B \cite{pratap2023scaling}. MMS-1B also has a version that is already fine-tuned to ASR of 1162 languages (MMS-1B-all), and language-specific adapter models. These adapters can be used to derive an optimized, low-rank language-specific fine-tuned model for each individual language.  We also experimented with fine-tuning this model for Bengali. Instead of fine-tuning adapter modules that are a common strategy with MMS-1B-all, we discarded the adapter modules and fine-tuned all layers of this model, using 3 epochs, effective batch size of 12 utterances and a learning rate of 3e-5.

The comparative results on the Bengali development data, produced with shallow fusion using a 4-gram LM trained on speech transcripts, are presented in Table \ref{tab:backbone-comp}. All models were trained with the same hyperparameters (except for MMS-1B-all) and no data augmentation was applied. As illustrated by the data, the MMS-1B model and its ASR-fine-tuned version achieved the lowest WER. Therefore, our efforts focused on the fine-tuning of those particular models.

\subsection{Final system}

For both languages, our final Track 3 system was based on four sub-models: (1) the Conformer wav2vec2.0 model from Track 1, (2) MMS-1B backbone model finetuned to the target language, (3), MMS-1B-all ASR model finetuned to the target language and (4) MMS-1B-all ASR model finetuned on the combined Bengali and Bhojpuri data. All models were trained with ADA and adapted to dialects using prefix tuning. Results of the official baseline model (codenamed fairseq-wav2vec2), our four individual models and their fusion are listed in Table \ref{tab:track3-results}. Our submission resulted in the lowest WER scores for both languages across all challenge tracks.

\begin{table}[tb]
\caption{Track 3 WERs of our individual systems (after optimizing LM weight) and their fusion, along with the official challenge baseline WERs.}
\label{tab:track3-results}
\begin{tabular}{l|cc|cc}
\toprule
                                 & \multicolumn{2}{c|}{Bengali} & \multicolumn{2}{c}{Bhojpuri} \\ \midrule
                              & Dev          & Test         & Dev           & Test         \\ \midrule 
Challenge baseline               & 18.1         & 20.7         & 19.1          & 18.7         \\ \midrule \midrule
Track 1 Conformer                & 15.2         &              & 15.5          &              \\
MMS-1B                           & 14.5         &              & 14.6          &              \\
MMS-1B-all                       & 13.4         &              & 13.7          &              \\
MMS-1B-all, bilingual ft. & 13.2         &              & 13.6          &              \\ \midrule
\textbf{Fusion}                           & \textbf{12.4}         & \textbf{15.1}         & \textbf{13.0}          & \textbf{13.9}   \\ \bottomrule
\end{tabular}
\end{table}

\subsection{Post-evaluation experiments: TTS-based data augmentation}

After the submission deadline, we explored the use of text-to-speech (TTS) models for data augmentation. The goal was to generate synthetic speech data that could be used to augment the limited training data and improve the performance of our ASR system. This method has been successfully used in several low-resource ASR scenarios \cite{li2018training, rosenberg2019speech, zhong22_interspeech}. Due to time constraints, we could only experiment with Bengali data. 

For the generation of synthetic speech, we utilized the VITS TTS model \cite{kim2021conditional}.  VITS can be defined as conditional variational inference augmented with normalizing flows and an adversarial training process. It has a reputation for creating natural-sounding audio with diverse rhythms corresponding to the input text.

To make the TTS model language-specific, we first generated phonemes for the Bengali language using a grapheme-to-phoneme (G2P) phonemizer \cite{bernard2021phonemizer}. The underlying technology for the G2P phonemizer was the \textit{espeak-ng} backend, which supports over 100 languages, including Bengali. These phonemes, along with their matching audio files, were input into the VITS model for training. Finally, the VITS model underwent training for 11 epochs with a batch size of 64.

We then used this TTS model to augment our speech data. New sentences were produced through random sampling from a 4-gram model,  trained on the speech transcripts. Each sentence was then transformed into a synthetic speech waveform via the VITS model for a randomly chosen speaker from the training set. This freshly generated speech data was merged with the original training data, effectively broadening the data set available for training our ASR system.

We evaluated the impact of different augmentation strategies by fine-tuning MMS-1B model (1) on the original training data, (2) original data augmented with TTS data, (3) ADA, and  (4) ADA combined with TTS data. Results (without any dialect adaptation) are listed in Table \ref{tab:tts}. Decoding was performed with no LM, with a (weak) LM trained on only speech transcripts, and with an optimized interpolated LM described in Section \ref{sec:lm}.
It can be seen that TTS-based augmentation actually deteriorates WER results, while ADA combined with TTS gives the best results across all experiments. Interestingly, the large improvement from ADA alone vanishes when using shallow fusion with LM. This can be explained by the fact that LM effectively acts as a regularization mechanism for models that have overfit on the limited text transcripts of the training data.

Due to time constraints imposed by the tight schedule of the challenge, we were not able to apply
dialect adaptation and estimate the contribution of TTS-based augmentation to the
full system.

\begin{table}[tb]
\caption{Bengali WER results with MMS-1B models finetuned with different data augmentation strategies.}
\centering
\label{tab:tts}
\begin{tabular}{l|cc}
\toprule
Augmentation              & No LM & With LM \\ \midrule
No augmentation &    19.3  &  17.4 / 14.7   \\
TTS          &      19.5 &  18.4 / 15.1 \\
ADA         &  18.0   &  17.4 / 15.0  \\
TTS+ADA       &   17.5 &  16.7 / 14.4  \\ 
\bottomrule
\end{tabular}
\end{table}

\section{Conclusion}

This paper described TalTech's systems that obtained the lowest WERs in Tracks 1 and 3 of the ASRU 2023 MADASR Challenge across all participating teams. Our approach was largely based on pretraining and fine-tuning wav2vec2.0 models. The key components of our systems were word-replacement based data augmentation, dialect-adaptive prefix tuning and efficient fusion of individual models. Post-evaluation experiments suggested that the models can further benefit from TTS-based data augmentation.

\section{Acknowledgments}

Authors acknowledge the TalTech supercomputing resources made available for conducting this research.

\bibliographystyle{IEEEbib}
\bibliography{refs}

\end{document}